\documentclass[11pt]{article}

\usepackage[preprint]{acl}

\usepackage{times}
\usepackage{latexsym}

\usepackage[T1]{fontenc}

\usepackage[utf8]{inputenc}

\usepackage{microtype}

\usepackage{inconsolata}

\usepackage{graphicx}

\usepackage{hyperref}
\usepackage{url}
\usepackage{booktabs}
\usepackage{multirow}
\usepackage{xcolor}
\usepackage{lineno}
\usepackage{amsmath}
\usepackage{xspace}
\usepackage{tikz}
\usepackage{subcaption}
\usetikzlibrary{arrows.meta, positioning, shapes.geometric}

\definecolor{darkblue}{rgb}{0, 0, 0.5}
\definecolor{pipeblue}{RGB}{52, 120, 190}
\definecolor{pipegray}{RGB}{230, 230, 235}
\hypersetup{colorlinks=true, citecolor=darkblue, linkcolor=darkblue, urlcolor=darkblue}

\usepackage{multirow}
\usepackage{enumitem}
\usepackage{adjustbox}
\usepackage{amsmath}
\usepackage{booktabs}
\usepackage{array}
\usepackage{longtable}
\usepackage{amssymb}
\usepackage{soul}
\usepackage{siunitx}
\usepackage{xcolor}
\usepackage{graphicx}
\usepackage{booktabs}
\usepackage{graphicx}
\usepackage[table]{xcolor}
\usepackage{tabularx}

\definecolor{bestrow}{RGB}{232,245,233}
\definecolor{groupgray}{RGB}{245,245,245}

\newcommand{\camerareadytext}[1]{}

\usepackage{listings}
\usepackage{xcolor}
\definecolor{tal-color}{HTML}{12b33d}

\title{Re-Centering Humans in LLM Personalization}

\author{
  Lechen Zhang$^{\dagger}$ ~~~~~~ Jiarui Liu$^{\sharp}$ ~~~~~~ \textbf{Tal August}$^{\dagger}$ \\
$^\dagger$University of Illinois Urbana-Champaign  ~~~~
$^\sharp$Carnegie Mellon University  \\
{ \tt \{lechenz3, taugust\}@illinois.edu}
~~~~ { \tt jiaruil5@andrew.cmu.edu}
}

\begin{document}
\maketitle

\begin{abstract}

Despite growing interest, most evaluations of large language models' (LLMs') personalization abilities have relied on synthetic data. It remains unclear how well current personalization systems work for real users. In this paper, we study the gap in LLM personalization performance when using synthetic versus human data. We collect human conversations (550 conversations) and judgments across three stages of personalization: extracting user attributes from conversations (5,949 judgments), pairing relevant attributes with new prompts (11,919), and incorporating relevant attributes into a personalized response (1,101). Incorporating human data reveals system limitations at each stage. Models struggle to extract attributes from human conversations, disagree with human judgments on relevant attributes, and generate personalized responses that humans judge no better than generic responses (though that LLM judges widely rate as better). We introduce two lightweight training-based interventions that shift automated personalization evaluation closer to human data in our first two stages. However, in our third stage we find that learned reward models achieve only modest correlation with human ratings, suggesting that human-aligned personalization quality judgments are difficult to model directly. Our collected data provides a foundation for studying how models should extract, select, and incorporate user information in ways that humans find useful.

\end{abstract}

\section{Introduction}
\label{sec:intro}

Large language models' (LLMs') ability to personalize responses to different users is becoming increasingly sought-after as models are deployed to more people across more domains \citep{shaikh2025creating}. Despite growing interest, it remains unclear how well current personalization systems work for real users. Past efforts evaluating models' personalization abilities have predominantly relied on synthetic data: user personas \citep{personamem}, simulated conversations \citep{kim2025cupid}, and LLM-based evaluations \citep{zhao-etal-2025-personalens}. While synthetic data enables large-scale experimentation, synthetic responses deviate significantly from human ones \citep{naous2026flipping, mehri2026measuringmitigatingdistributionalgap}. It is not clear if model performance on synthetic data maps neatly to real human experience. This tension highlights the central difficulty of personalization: the target of personalization is a human user, whose attributes and judgments on system effectiveness are often implicit, noisy, incomplete, and context-dependent \citep{10.1109/ICDM.2008.22, shaikh2025creating}. 

In this paper, we study the gap between synthetic and human data for evaluating LLM personalization. We frame personalization as a three-stage pipeline based on prior evaluations: models must (1) infer stable user attributes from prior conversations (user attribute extraction), (2) select the attributes relevant to a specific response (attribute relevance matching), and (3) incorporate relevant attributes to produce a response that improves over a generic one (personalized response generation). For each stage of our pipeline, we compare human and synthetic data. Our dataset contains 50 real users and 550 conversations drawn from existing conversational datasets, paired with human judgments for all three stages: 5,949 judgments on extracted user attributes, 11,919 on attribute--prompt pairings, and 1,101 on response preference judgments for personalized generation. Using our dataset, we compare model performance and alignment with human data and judgments.  

We find that incorporating human data reveals model limitations at each stage of personalization. Models struggle to accurately extract attributes from human conversations compared to synthetic ones: an additional 22\% of extracted user attributes from real conversations are judged problematic. Once attributes are extracted, models again face difficulties in pairing attributes with prompts: despite strong inter-LLM agreement, LLMs are misaligned with humans on relevance matching, over-identifying 20--40\% more attributes as relevant. Finally, even when relevant attributes are provided, humans disagree with models on what constitutes an effective personalized response: as generators, LLMs produce personalized responses that humans judge as no better than generic ones in 54.6\% of cases, while as judges, LLMs assign inflated scores that remain poorly aligned with human preferences.

We introduce two lightweight interventions that shift automated personalization evaluation closer to human data in our first two stages. For attribute extraction, a lightweight RoBERTa~\citep{liu2020roberta} verifier trained on human annotations can serve as an effective safeguard before extracted attributes are used downstream. For relevance matching, training-based methods, including supervised classification and GRPO~\citep{grpo}, reduce over-selection and better align relevance judgments with humans. For our third stage of personalized response generation, trained reward models achieve modest correlation with human ratings, suggesting that human-aligned personalization quality judgments are difficult to model directly.

Overall, our results show that personalization presents unique difficulties for automated and synthetic evaluations. We hope our collected data enables deeper evaluation of personalization systems and provides a foundation for studying how models should extract, select, and incorporate user information in ways that humans find useful. The code and dataset are available at \url{https://github.com/orange0629/recenter-personalization}.

\begin{figure*}[ht]
    \centering
    \includegraphics[width=0.9\textwidth]{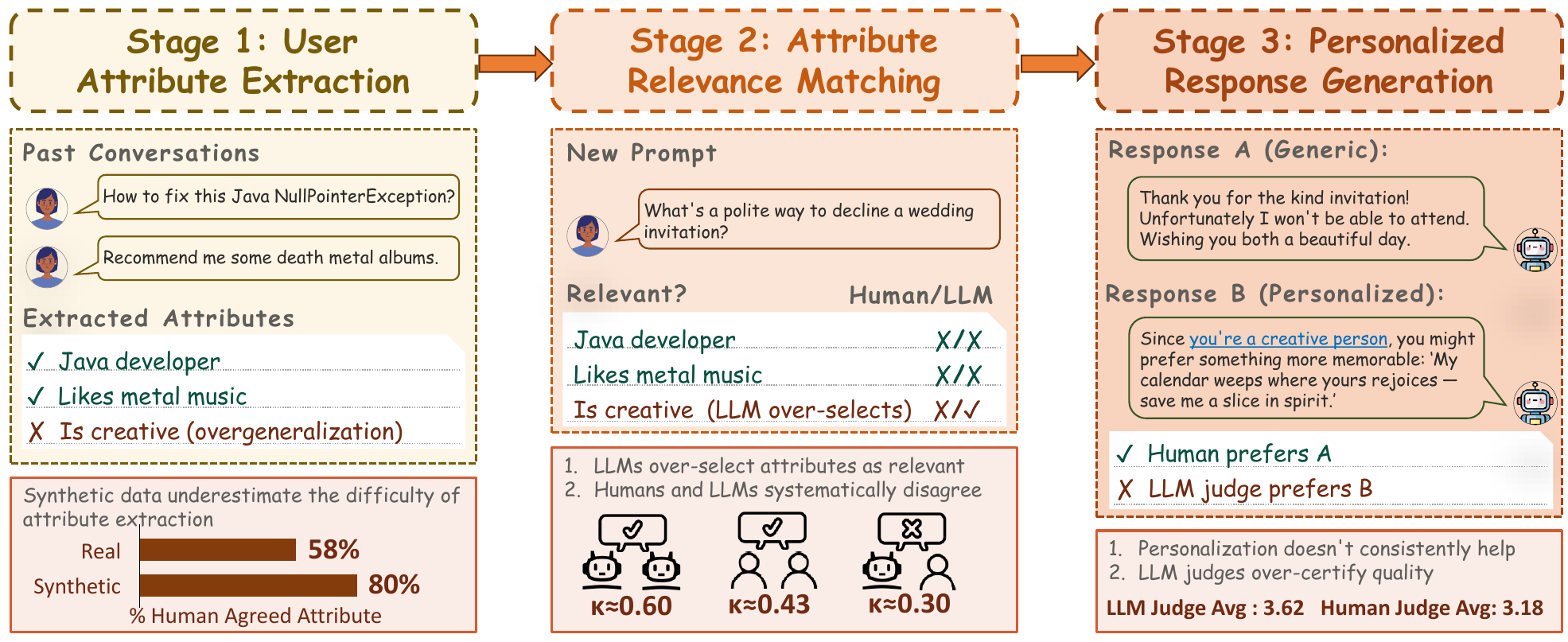}
    \caption{Overview of our three-stage personalization framework. \textbf{(1) User Attribute Extraction} from conversation history, \textbf{(2) Attribute Relevance Matching} for the current interaction context, and \textbf{(3) Personalized Response Generation} based on the selected attributes. By incorporating human-grounded data, we reveal limitations at each stage of personalization: models struggle to extract accurate attributes from human conversations, pair those attributes with new prompts, and determine what makes a personalized response effective.}
    \label{fig:pipeline}
    \vspace{-9pt}
\end{figure*}

\vspace{-3pt}
\section{A Three-Stage Framework for LLM Personalization}
\vspace{-3pt}
\label{sec:pipeline}
LLM personalization spans a spectrum from monolithic to decomposed approaches. End-to-end benchmarks evaluate personalization quality as a single black-box task, making it difficult to diagnose where a system fails \citep{zollo2025personalllm,zhao-etal-2025-personalens,kim2025cupid,zhao2025do}. A parallel line of work argues for more explicit decomposition \citep{wang2023cue,zhuang2024hydra,shi2025retrieval,xu2026toward,du2026optimizing, mehri2026multisessioncollablearninguserpreferences}, showing that factorizing personalization into stages such as retrieval, ranking, and personalized generation yields consistent gains over monolithic approaches.

We frame LLM personalization as a three-stage pipeline (Figure~\ref{fig:pipeline}): \textbf{(1) user attribute extraction} from conversation history, 
\textbf{(2) attribute relevance matching} for the current interaction context, and 
\textbf{(3) personalized response generation} based on the selected attributes. This framework is based on prior personalization efforts \citep[e.g., relevance matching is inspired by structured memory lookup,][]{salemi-etal-2024-lamp}, though we narrow our focus to initial response personalization (i.e., given a user and their previous conversations, personalize the next response), and leave subsequent personalization stages, such as updating outdated memories~\citep{locomo}, to future work. 

\paragraph{Stage 1 --- Attribute extraction.}

We define a \textbf{user attribute} as a stable, long-term, and context-independent statement about the user, including both user preferences and profile information (e.g., \emph{``user prefers bullet-points''}, or \emph{``user is a Java developer''}). We focus on attributes that are unlikely to change within a single conversation as an initial step to formalizing our pipeline. This definition follows a common practice in personalization and memory systems, where past interactions are synthesized into persistent user profiles (similar to personas) that can later be retrieved for personalized generation~\citep{memorybank, kang-etal-2025-memory}.

\paragraph{Stage 2 --- Relevance matching.}
We define \textbf{attribute relevance} as whether a user attribute should influence the model's response to a specific prompt.\camerareadytext{ Not all known attributes should be applied to every response: a user's preference for technical depth may be useful for a programming question but distracting when they seek emotional support.} Prior systems often approximate this stage as a retrieval problem, selecting user information through semantic similarity or structured memory lookup~\cite{salemi-etal-2024-lamp, wu2025longmemeval, sun-etal-2025-persona}. However, semantic similarity does not necessarily imply personalization relevance, and lexically distant attributes may still affect how a response should be framed~\citep{okite2026lucidredefiningrelevancelifelong}. We therefore treat relevance selection as a reasoning problem, where the system must decide which attributes would meaningfully improve the response, which should remain unused, and how \camerareadytext{each selected }attributes should affect the response without introducing irrelevant or excessive personalization.

\paragraph{Stage 3 --- Personalized response generation.}

Once relevant attributes have been selected, the model must determine \emph{how} to use them in the final response. We define successful \textbf{personalized generation} as producing a response that improves over a generic response. Existing work often acknowledges related concerns, but rarely directly studies the quality of this stage. Instead, most work uses adjacent metrics such as task performance, preference alignment, or profile consistency~\citep{salemi-etal-2024-lamp, personamem}.
We therefore evaluate not only whether selected attributes affect the output, but also whether they are used in a contextually appropriate and user-acceptable manner.

\section{Re-Centering Humans in the Personalization Pipeline} 
\label{sec:problems}

Although humans are the subjects of personalization, most existing work proxies the role of human input with synthetic personas, simulated users, synthetic dialogues, or LLM-based judges \cite{salemi-etal-2024-lamp, personamem, liu-etal-2025-synthetic, zhao-etal-2025-personalens}. While these designs enable large-scale evaluation, it is not clear that current approaches can faithfully simulate human users~\cite{ivey2024realroboticassessingllms, dong-etal-2024-llm, naous2026flipping, mehri2026measuringmitigatingdistributionalgap}. 
As a result, existing personalization benchmarks may inject systematic bias or overestimate system capabilities. We argue that progress on LLM personalization requires re-centering humans in the personalization pipeline by: (1) grounding data to real human interactions (\S\ref{sec:diversity}), and (2)  judging personalization quality with humans (\S\ref{sec:relevance} and \S\ref{sec:behavior}). In each of the sections below, we illustrate how incorporating human conversations and judgments highlights current limitations in personalization evaluation.

\paragraph{Models}

In this study, we mainly focus on five recent and widely-used open-weight and proprietary LLMs. 
The open-weight LLMs include Llama-3.3-70B~\citep{llama3}, Qwen3.5-27B~\citep{qwen2026qwen35}, and Gemma-4-31B~\citep{googledeepmind2026gemma4}; 
the proprietary LLMs include Claude-Sonnet-4.6 and GPT-5.4. These open-weight LLMs have shown strong quality-cost trade-offs, making them practical choices for large-scale personalization experiments.

\section{Grounding personalization to human interactions (Stage 1)}
\label{sec:diversity}
We investigate how attribute extraction (Stage 1) differs when using human or synthetic conversations. We find that human conversations contain richer and more diverse attributes (\S\ref{sec:attributeDiversity}), and that extracting attributes from human conversations is more error-prone than synthetic ones (\S\ref{sec:attributeNoise}). 

\subsection{Experiment}

\paragraph{Data} We use WildChat~\citep{zhao2024wildchat} as the base source for human conversations and cluster multiple conversations by user IP\footnote{The use of hashed IP addresses as approximate user identifiers follows prior work on WildChat~\citep{zhao2024wildchat, naous2026flipping}.}, resulting in 98,334 users. We filter to only English conversations from active and authentic users,\footnote{We filter users with enough (at least 3 conversations and 15 turns) interaction history and with no signs of automated or malicious usage (LLM judged). Full preprocessing details are in Appendix~\ref{app:wildchat}.} yielding 16,573 users in total. For synthetic data, we include three recent personalization datasets: CUPID~\citep{kim2025cupid}, PrefEval~\citep{zhao2025do}, and PersonaLens~\citep{zhao-etal-2025-personalens}. 

\paragraph{Experiment Setting} To extract user attributes, we split each user history into shorter chunks that fit within all LLMs' context window (50K tokens) and prompt an LLM to list user-related attributes expressed or implied in each chunk, along with a reasoning and confidence score for each extracted attribute. We use Llama-3.3-70B as our backbone model for our attribute extraction experiments.\footnote{We found Llama-3.3-70B to be a competitive choice for this task, as it performed best among the open-source models we tested based on manual inspection. We do not use closed-source models because some extremely long conversations can make large-scale extraction prohibitively expensive.} We filter out attributes with confidence scores below 0.4, and use an agglomerative clustering algorithm to merge overlapping attributes. Full implementation details are in Appendix~\ref{app:attr-extraction}.

\paragraph{Human Annotation} We ask annotators to evaluate extracted attribute quality by labeling each attribute as accepted, uncertain, or rejected. We sampled 77 users in total, including 47 WildChat users and 10 users from each of the three synthetic datasets for the annotation task. We recruit annotators in two rounds. First, we launch a pilot study with 250 attributes from 10 users, and recruit five annotators on Prolific\footnote{\url{https://www.prolific.com/}} who are fluent in English and met standard platform quality criteria. We assess annotation quality by checking each annotator's label distribution and inter-annotator agreement, with author inspection for potential outliers. We find that four of the five annotators produce reasonable annotations, with an average Cohen's $\kappa$ of around 0.35. We then recruit three of these annotators for the full study, covering 1,983 attributes from 77 users. The recruited annotators achieve a Cohen's $\kappa$ of 0.314. More details are in Appendix~\ref{app:human-annot}.

\begin{table}[t]
\centering
\small
\setlength{\tabcolsep}{6pt}
\renewcommand{\arraystretch}{1.12}

\resizebox{\linewidth}{!}{%
\begin{tabular}{lc}
\toprule
\textbf{Dataset} & \textbf{Avg. Inter-user Dist.} $\uparrow$ \\
\midrule

\rowcolor{gray!12}
\multicolumn{2}{l}{\textit{Synthetic baselines}} \\
\quad PrefEval~\citep{zhao2025do} & 0.689 \\
\quad CUPID~\citep{kim2025cupid} & 0.649 \\
\quad PersonaLens~\citep{zhao-etal-2025-personalens} & 0.706 \\

\addlinespace[2pt]
\rowcolor{gray!12}
\multicolumn{2}{l}{\textit{Real data}} \\
\quad WildChat~\citep{zhao2024wildchat} & 0.656 \\
\rowcolor{green!10}
\quad \textbf{WildChat (Filtered)} & \textbf{0.737} \\

\bottomrule
\end{tabular}%
}

\caption{
Diversity of different datasets, calculated by average pairwise inter-user cosine distance among the sentence embeddings of all extracted user attributes. While unfiltered real data does not exhibit a clear diversity advantage, a simple preprocessing yields substantially higher diversity than all synthetic baselines.
}
\label{tab:diversity}
\vspace{-8pt}
\end{table}

\subsection{Results}

\paragraph{Real users can be more diverse than synthetic alternatives}
\label{sec:attributeDiversity}
A central goal of many synthetic conversation datasets is to approximate the diversity of real user data~\citep{personamem}. In our study, we define diversity as the average pairwise inter-user cosine distance among the sentence embeddings~\citep[\texttt{all-MiniLM-L6-v2};][]{reimers-gurevych-2019-sentence} of all extracted user attributes, where higher values indicate greater variation across users. However, our experiments reveal that \textbf{uniformly sampling users from WildChat does not necessarily produce higher diversity than recent synthetic datasets} (Table~\ref{tab:diversity}). This is partly because WildChat contains many homogeneous users (e.g., a large proportion of users only talk about coding), and many users do not exhibit deep model usage (e.g., short-term LLM users). 

Nevertheless, \textbf{real data provides a large enough pool from which a more diverse subset can be selected}. We obtain a subset of 5,000 WildChat users that is substantially more diverse than all synthetic datasets we compare against (highlighted in green in Table~\ref{tab:diversity}) by applying basic filtering and diversity-based sampling (details in Appendix~\ref{app:diversity-sampling}, examples of sampled users in Appendix Table~\ref{tab:wildchat-user-examples}).

\begin{figure}[tb!]
    \centering
    \includegraphics[width=0.95\columnwidth]{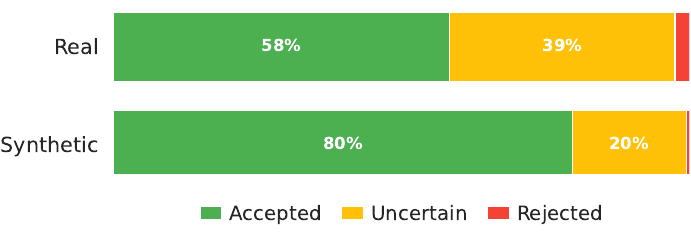}
        \caption{Human annotation of extracted user attributes from synthetic and real conversations, categorized into \colorbox{green!40}{accepted}, \colorbox{yellow!50}{uncertain}, or \colorbox{red!40}{rejected}. Compared with synthetic datasets, real conversations yield a lower acceptance rate and higher proportions of uncertain and rejected judgments, indicating that reliable attribute extraction is more challenging on real data. %
}
    \label{fig:task1_annotation}
    \vspace{-8pt}
\end{figure}

\paragraph{Real conversations make attribute extraction more challenging}
\label{sec:attributeNoise}

Human conversations are often noisier than synthetic equivalents \citep{naous2026flipping}, will this impact a system's ability to reliably extract attributes? We use the 1,983 human annotated attributes of 77 users from three annotators described above to compare attribute validity across real and synthetic conversations. As shown in Figure~\ref{fig:task1_annotation}, attributes extracted from human conversations contain substantially higher proportions of uncertain and rejected cases than those extracted from the synthetic benchmarks. This suggests that \textbf{extracting attributes that humans find valid is considerably more challenging from real conversations, and that synthetic data may underestimate the difficulty of this stage.}

To analyze these disagreements, we use GPT-5.4 to categorize the reasons behind all $1{,}225$ uncertain attributes by annotators (details in Appendix~\ref{app:attr-uncertainty}). The most common failure mode is \textit{overgeneralization} (53.9\%), where narrow evidence is extrapolated into a stable user trait, such as inferring that a user ``is learning French'' from one translation request. Other common cases include \textit{missing evidence} (20.3\%), where the attribute is not supported by the visible excerpt, and \textit{task-context confusion} (16.1\%), where task content is mistaken for user information, such as inferring that the user ``has five years of marketing experience'' from a fictional cover-letter prompt.

\subsection{Analysis: Attribute Verification as a Lightweight Safeguard}
\label{sec:better-verification}

To address the attribute extraction challenge posed by human conversational data, we propose a simple post-extraction verification and optimization step. Instead of directly treating extracted attributes as reliable user profiles, personalization systems can first verify whether each attribute is sufficiently supported by the conversation history, then ask the extractor to revise or remove unsupported attributes before downstream personalization. We evaluate three verifiers: (1) zero-shot LLM prompting with the same instruction given to annotators (Appendix~\ref{prompt:attr-verifier-basic}), (2) LLM prompting with an optimized prompt (Appendix~\ref{prompt:attr-verifier-optimized}) based on our error analysis in \S\ref{sec:attributeNoise}, and (3) a RoBERTa classifier trained on our human annotations (training details in Appendix~\ref{app:roberta-task4}).
\begin{table}[t]
\centering
\small
\setlength{\tabcolsep}{5pt}
\renewcommand{\arraystretch}{1.12}
\resizebox{0.9\linewidth}{!}{%
\begin{tabular}{lcccc}
\toprule
\textbf{Verifier} & \textbf{Accuracy} & \textbf{Precision} & \textbf{Recall} & \textbf{F1} \\
\midrule

\rowcolor{gray!10}
Llama-3.3-70B & 0.655 & 0.531 & 0.233 & 0.324 \\
\quad + optimized prompt & 0.681 & 0.564 & 0.425 & 0.484 \\

\addlinespace[1pt]
\rowcolor{gray!10}
Qwen3.5-27B & 0.642 & 0.478 & 0.471 & 0.475 \\
\quad + optimized prompt & 0.628 & 0.476 & 0.758 & 0.585 \\

\addlinespace[1pt]
\rowcolor{gray!10}
Gemma-4-31B & 0.652 & 0.506 & 0.562 & 0.532 \\
\quad + optimized prompt & 0.643 & 0.496 & 0.877 & 0.634 \\

\addlinespace[1pt]
\rowcolor{gray!10}
GPT-5.4 & 0.676 & 0.547 & 0.479 & 0.511 \\
\quad + optimized prompt & 0.589 & 0.461 & \textbf{0.973} & 0.626 \\

\addlinespace[1pt]
\rowcolor{gray!10}
Claude-Sonnet-4.6 & 0.667 & 0.525 & 0.575 & 0.549 \\
\quad + optimized prompt & 0.614 & 0.476 & 0.932 & 0.630 \\

\addlinespace[2pt]
\rowcolor{green!10}
\textbf{RoBERTa (Trained)} & \textbf{0.748} & \textbf{0.591} & 0.926 & \textbf{0.726} \\

\bottomrule
\end{tabular}%
}
\caption{Performance of attribute verifiers for detecting problematic extracted attributes. The optimized prompt substantially improves recall for most LLMs. The supervised RoBERTa verifier achieves the best F1, suggesting that human annotations help calibrate verification toward human standards of evidential support.}
\label{tab:attribute-verifier}
\vspace{-10pt}
\end{table}

Table~\ref{tab:attribute-verifier} shows three key findings. First, \textbf{optimized verification prompts based on error analysis substantially improve the verifier's ability to identify unsupported attributes}. This is useful because recall is especially important for verification: unsupported attributes missed by the verifier will be passed downstream as reliable user profiles, whereas supported attributes incorrectly flagged by the verifier can still be preserved or softened during the subsequent refinement step. The optimized prompt significantly improves recall for most models, suggesting that explicit instructions about over-generalization and insufficient evidence help verifiers catch more problematic attributes. Second, \textbf{LLM verifiers exhibit clear cross-model differences after prompt optimization}. Before optimization, most models perform relatively similarly, but after optimization, stronger closed-source models such as GPT-5.4 and Claude achieve much higher recall, while Llama-3.3-70B remains weak; Gemma-4-31B reaches an F1 score close to GPT-5.4 and Claude, but with lower recall. Third, \textbf{a supervised RoBERTa verifier provides the strongest practical trade-off between precision and recall}. It achieves the highest overall F1 while maintaining strong recall, suggesting that a small verifier trained on human annotations can provide a reliable and lightweight safeguard before extracted attributes are used for personalization.

We further use the RoBERTa verifier as a reflection signal for attribute optimization. Attributes flagged as unsupported are sent back to the extraction model for revision. In a small-scale study on 250 attributes, this step increases the average human acceptance rate from 58\% to above 90\%. This suggests that many extraction errors are recoverable with a simple verify-and-refine step.

\section{Aligning Personalized Relevance Selection with Human (Stage 2)}
\label{sec:relevance}

Once attributes have been extracted (Stage 1), personalization requires selecting which attributes are relevant to a specific prompt. Here we test humans' and LLMs' agreement on relevance judgments.

\subsection{Experiments}

\paragraph{Data}
We use LIMA~\citep{zhou2023lima}, a high-quality human instruction dataset for SFT, as the source of dialogue prompts to pair with our extracted attributes. We randomly sample 41 prompts and pair them with the 47 WildChat users whose attributes were annotated in Stage 1. Rather than pairing prompts and users completely at random, we first use Llama-3.3-70B as a coarse filter over all possible prompt--user combinations, and remove combinations with fewer than two potentially relevant attributes. We then sample 4 users for each prompt from the retained pairs. The selected attributes (3,969 in total) are then shuffled and presented to human annotators and LLM judges for relevance annotation.

\begin{figure}[tb!]
    \centering
    \includegraphics[width=0.95\columnwidth]{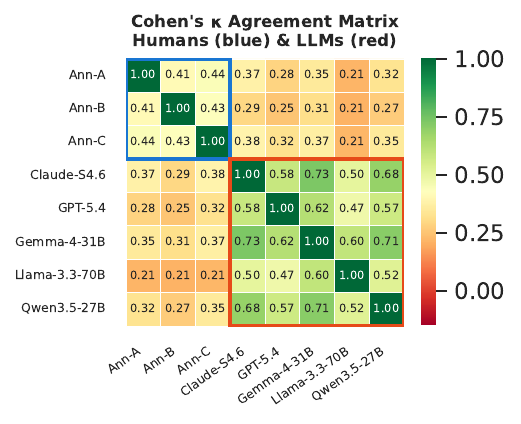}
    \caption{Full pairwise Cohen's $\kappa$ between all three human annotators and five LLMs. Human--human agreement (top-left block) is consistently higher
    than LLM--human agreement (off-diagonal), confirming that LLMs are not reliable proxies for human relevance judgment.}
    \label{fig:kappa_full}
    \vspace{-8pt}
\end{figure}

\paragraph{Experiment Setting}
We formulate relevance selection as a binary judgment task and compare human and LLM judges. Given a prompt and a candidate user attribute, both humans and LLMs are asked to decide whether the attribute should be considered when personalizing the response. An attribute is labeled relevant if it would affect the response in any way (either explicitly mentioned or implicitly impacted) and irrelevant only if the response would be unchanged without it. We collect annotations from three human annotators (details below) and five LLMs (\S\ref{sec:problems}).

\paragraph{Human Annotation}

We collect human relevance judgments in two rounds. In the pilot round, we invite seven annotators to label 631 attribute--prompt pairs. We assess annotation quality by their yes/no label distribution and pairwise agreement, and remove two annotators with distributions far from the other annotators after manual inspection of their responses. We then invite three of the annotators, whose pairwise Cohen's $\kappa$ falls in the 0.4--0.6 range, to complete the full study of 3,969 attribute--prompt pairs from 41 prompts. More details and the annotation interface are in Appendix~\ref{app:human-annot}.

\subsection{Results}

\paragraph{Humans and LLMs systematically disagree on relevance selection} Humans and LLMs each show substantial internal agreement but disagree with each other. As shown in Figure~\ref{fig:kappa_full}, the three human annotators achieve an average pairwise Cohen's $\kappa$ of 0.426, which is substantial given the subjective nature of the task \citep{da-san-martino-etal-2019-fine, landis1977measurement}. Surprisingly, LLMs agree even more strongly with one another, reaching an average $\kappa$ of 0.597 across diverse open-source and closed-source models. However, LLM--human agreement is much lower, with an average $\kappa$ of only 0.300. One clear source of this misalignment is that LLMs assign substantially higher relevance rates: each human annotator marks only around 20\% of attributes as relevant, whereas LLMs mark 40--60\% as relevant across models. This pattern suggests that both humans and models form internally consistent judgments, but LLMs' decision boundaries are substantially misaligned with humans, making them unreliable proxies for human relevance judgment.

\begin{figure}[tb!]
    \centering
    \includegraphics[width=\columnwidth]{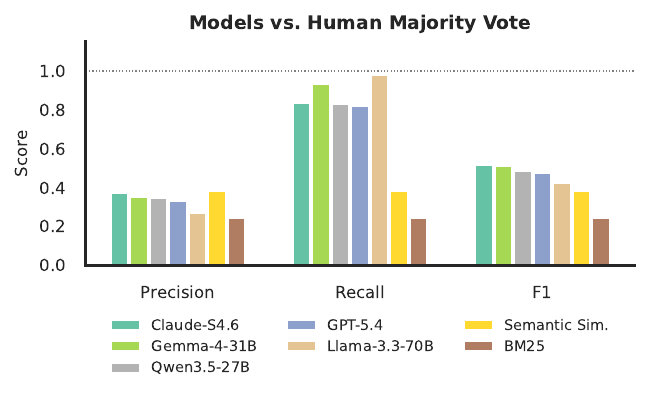}
        \caption{Precision/Recall/F1 of attribute relevance selection against human majority vote. Retrieval-based methods perform poorly, indicating that lexical similarity are insufficient. Although LLM judges outperform these baselines, their limited precisions show that they remain unreliable, with high recall largely resulting from over-selection.}
    \label{fig:task5_annotation}
\end{figure}

\paragraph{Attribute relevance cannot be reduced to semantic similarity.} As discussed in \S\ref{sec:pipeline}, most existing work treats relevance selection as a retrieval problem or constructs personalization data based on semantic relatedness, which can underestimate the difficulty of this stage. To test this, we compare BM25~\citep{bm25} and sentence-embedding similarity as retrieval-based baselines, and use the majority vote\footnote{We use majority vote as a human-consensus label, following common practice in crowdsourced annotation~\citep{snow-etal-2008-cheap,10.1162/tacl_a_00449}; this label achieves Cohen's $\kappa > 0.7$ with each individual annotator.} of the three human annotators as the ground truth.  Figure~\ref{fig:task5_annotation} reports the F1 score of each method. BM25 and semantic similarity achieve only 0.243 and 0.384 respectively, both substantially below the performance of LLM judges. This suggests semantic relatedness is insufficient for connecting attributes with a response.

Among LLM judges, recall is consistently high, indicating that models tend to over-select attributes as relevant. However, their precision remains low, with all models below 0.4. Claude-S3.7, Gemma4, and Qwen3.5 achieve the top three F1 scores, while GPT-5.4 surprisingly falls behind these smaller open-source models. This aligns with our qualitative observations: GPT-5.4 tends to personalize on a broad range of attributes, similar to Llama3.3-70B, another low-performing model.

\subsection{Analysis: Aligning Attribute Relevance Selection with Better Reasoning}
\label{sec:better-relevance}

To address the relevance selection challenge identified in Section~\ref{sec:relevance}, we explore whether models can be better aligned with human judgments on which attributes should actually influence a response. As shown earlier, more than 60\% of the attributes that LLMs mark as relevant are not considered relevant by humans, which can introduce unnecessary or distracting personalization. We noticed that prompt refinement brings only marginal gains for this alignment problem. Therefore, we evaluate two training-based approaches: a RoBERTa classifier trained as a binary relevance predictor, and a Qwen3-4B reasoning model optimized with GRPO. Training details for both methods are in Appendix~\ref{app:training-task5}.

\begin{table}[t]
\centering
\small
\setlength{\tabcolsep}{5pt}
\renewcommand{\arraystretch}{1.12}
\resizebox{\linewidth}{!}{%
\begin{tabular}{lcccc}
\toprule
\textbf{Model} & \textbf{Accuracy} & \textbf{Precision} & \textbf{Recall} & \textbf{F1} \\
\midrule

Llama-3.3-70B & 0.529 & 0.272 & \textbf{0.979} & 0.426 \\
Qwen3.5-27B & 0.689 & 0.345 & 0.832 & 0.488 \\
Gemma-4-31B & 0.685 & 0.355 & 0.934 & 0.514 \\
GPT-5.4 & 0.676 & 0.334 & 0.819 & 0.474 \\
Claude-Sonnet-4.6 & 0.719 & 0.372 & 0.838 & 0.515 \\

\addlinespace[2pt]
\midrule
\rowcolor{green!10}
\textbf{RoBERTa} & 0.859 & 0.608 & 0.605 & 0.606 \\
Qwen3-4B & 0.530 & 0.267 & 0.942 & 0.417 \\
\rowcolor{green!10}
\quad \textbf{+ GRPO} & \textbf{0.870} & \textbf{0.611} & 0.674 & \textbf{0.641} \\

\bottomrule
\end{tabular}%
}
\caption{Performance on attribute relevance selection against human majority judgments. Zero-shot LLMs achieve high recall but low precision, indicating substantial over-selection of relevant attributes. Training-based methods substantially improve precision and accuracy, with RL-trained Qwen3-4B achieving the best F1.}
\label{tab:relevance-alignment}
\vspace{-8pt}
\end{table}

Table~\ref{tab:relevance-alignment} shows two main findings. First, \textbf{training-based alignment is more effective than zero-shot prompting}. Both RoBERTa and GRPO improve over all zero-shot LLMs in F1 and precision, suggesting that supervised training can better calibrate relevance selection toward human judgments. Second, \textbf{GRPO achieves the strongest performance among the methods we evaluate}. Compared with the base Qwen3-4B model, GRPO increases F1 and precision from 0.417/0.267 to 0.641/0.611, which is even higher than RoBERTa. This suggests that relevance selection may benefit from training LLMs to explicitly reason about whether an attribute should meaningfully affect the response.

We include an example in Appendix Table~\ref{tab:grpo-example} to show how GRPO changes the model's relevance reasoning. The untrained Qwen3-4B marks \emph{``the user is creative''} as relevant to a simple factual question, arguing that the response could \emph{``explain in a way that connects to creative processes.''} After GRPO, the model instead reasons, \emph{``Wait, the request asks for a basic definition, this attribute should not change the factual answer.''} This more cautious decision boundary better matches human judgments and reduces irrelevant attribute use before response generation.

\section{Human-Judged Personalization Quality (Stage 3)}
\label{sec:behavior}

We next investigate how humans and models judge personalization quality (i.e., how personalized a response is) given a set of relevant attributes.

\subsection{Experiments}

\paragraph{Experiment Setting} For Stage 3, we use the relevant attributes identified by human majority vote in Stage 2 as ground-truth inputs, and ask the five LLMs to generate responses conditioned on these attributes. For each instance, we present both LLM and human judges with two randomly shuffled responses: one generic (i.e., generated without personalization) and the other generated with ground-truth attributes. Annotators and LLMs are asked to judge which response they prefer on a five-point likert-style scale for each attribute, ranging from strongly preferring the non-personalized response to strongly preferring the personalized response.

\paragraph{Human Annotation} 
We invite the same three annotators from Stage 2 to judge personalized response quality. We measure inter-annotator agreement using both Spearman correlation and weighted Cohen's $\kappa$, which gives 0.325 and 0.310 respectively. Given the subjective nature of personalization preference, this indicates reasonable annotator consistency while still allowing for individual variation. More details about annotation are in Appendix~\ref{app:human-annot}.

\subsection{Results}

\begin{figure}[tb!]
    \centering
    \includegraphics[width=\columnwidth]{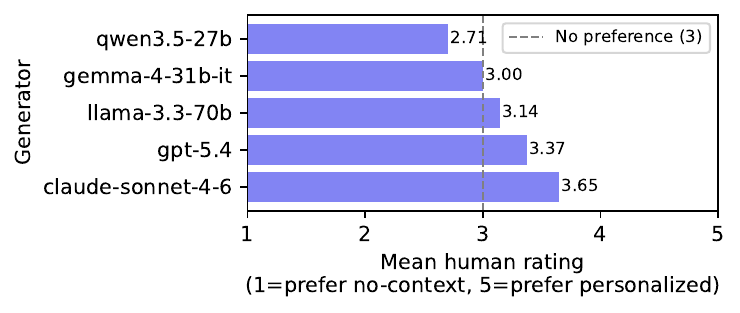}
        \caption{Mean human rating for each LLM generator. Human judgments show that personalized generation brings only marginal benefits to user experience even for recent models, and can even degrade response quality for some open-source models.}
    \label{fig:task6_annotation}
\end{figure}

\paragraph{Personalized responses do not consistently lead to improved response quality to users}
Based on collected human preference ratings on each LLM's generation, 54.6\% of personalized responses are judged by humans as no better than their generic counterparts (rating $\leq$ 3.0). As shown in Figure~\ref{fig:task6_annotation}, even the best-performing proprietary models' responses (GPT-5.4 and Claude-S4.6) are only slightly above the neutral baseline of 3.0, while open-source models like Qwen3.5 and Gemma-4 even degrade responses after personalization. This highlights the importance of cautious personalization: models must not only select relevant attributes accurately, but also decide carefully how to incorporate them into the generated response.

\begin{figure}[tb!]
    \centering
    \includegraphics[width=\columnwidth]{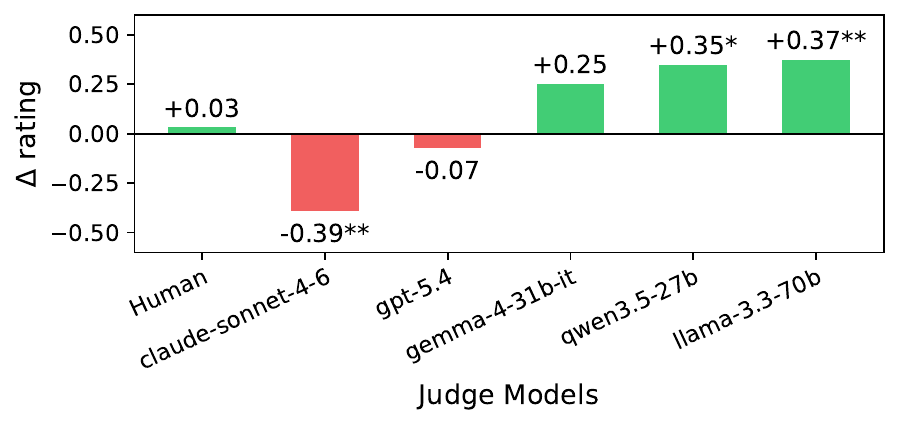}
        \caption{Each judge's sensitivity to explicit attribute invocation, measured as the mean rating gap between explicit-mention and no-mention responses. Open-source LLMs show a high sensitivity to explicit attribute mentions, while human judges and GPT-5.4 show little sensitivity, and Claude-S4.6 instead favors more implicit personalization. ($^{*}p<0.05$, $^{**}p<0.01$ for $\Delta$)}
    \label{fig:explicit_personalization_hack}
    \vspace{-6pt}
\end{figure}

\begin{figure}[tb!]
    \centering
    \includegraphics[width=\columnwidth]{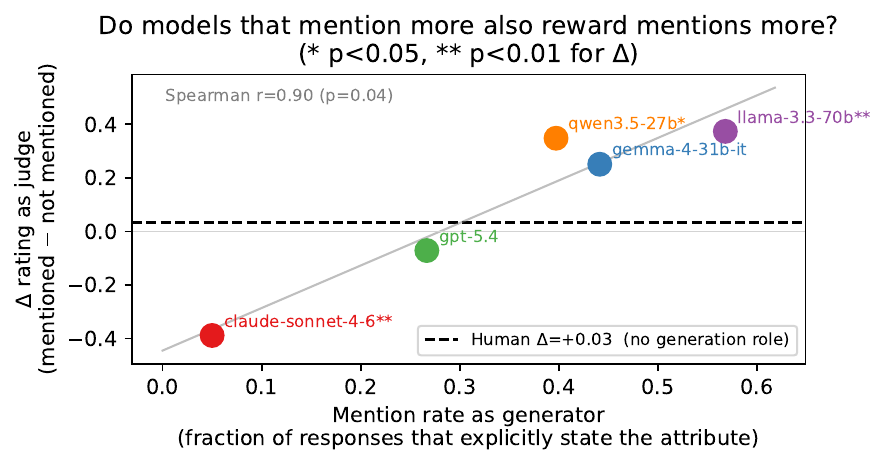}
        \caption{Relationship between a model's frequency to explicitly mention attributes when personalizing responses (x-axis) and its tendency to reward such mentions when judging responses (y-axis). The very high correlation suggests that models that mention attributes more often also tend to reward explicit mentions more strongly. Surface-level personalization behaviors may carry over from generation to evaluation.}
    \label{fig:explicit_personalization_hack_regression}
    \vspace{-6pt}
\end{figure}

\paragraph{LLM judges tend to overestimate personalization quality while only partially aligning with human preferences.} As shown in Table~\ref{tab:llm-judge-personalization}, humans assign lower ratings than all LLM judges, indicating that automated evaluation often overstates the benefit of personalization. Open-weight judges show very low Spearman correlations with human ratings, while proprietary judges perform better, with the strongest reaching about 0.37. \textbf{However, this modest correlation and systematic score inflation suggest that LLM judges remain unreliable substitutes for human evaluation.}

One reason is that LLM judges often over-reward visible personalization. A common failure is \emph{mechanical attribute invocation}, where a response explicitly mentions a user attribute, such as \emph{``Given your interest in machine learning...''}, without meaningfully adapting the content. Such responses may look personalized to LLM judges but feel robotic, presumptuous, or intrusive to users. 

To test this, we use GPT-5.4 to identify responses that explicitly mention relevant attributes and measure each judge's rating gap between explicit-mention and no-mention responses. As shown in Figure~\ref{fig:explicit_personalization_hack}, several open-weight judges assign higher scores to explicit mentions, while human annotators and GPT-5.4 show little sensitivity. Claude-S4.6 instead shows a significant negative gap, suggesting a preference for more implicit personalization.

We further ask whether models that explicitly mention user attributes more often as generators also reward such mentions more strongly as judges. As shown in Figure~\ref{fig:explicit_personalization_hack_regression}, the two behaviors are strongly correlated (Spearman $r=0.90$, $p=0.04$): models with higher explicit mention rates tend to assign larger rewards to explicit mentions during evaluation, suggesting that surface-level personalization preferences may transfer from generation to evaluation. Claude-S4.6 is a notable counter-directional case: it has the lowest explicit mention rate as a generator (only 5\%), and is the only judge that significantly penalizes explicit mentions. However, this penalty appears too strong relative to human judgments: human annotators show almost no gap between explicit-mention and no-mention responses ($\Delta=0.03$), while Claude-S4.6 gives explicit mentions much lower scores ($\Delta=-0.39$). GPT-5.4 is closer to humans on this dimension, with only a small negative gap ($\Delta=-0.07$). Overall, these results suggest that surface-level personalization behaviors may carry over from generation to evaluation, making LLM judges unreliable proxies for human preferences.

\subsection{Analysis: Learning Human-Aligned Personalization Rewards}
\label{sec:better-generation}

For judging personalized responses, a natural solution is to train a reward model that predicts human-aligned preference ratings. However, our results suggest that this stage remains challenging. We follow the common practice to train reward models~\citep{ouyang_gpt3} and evaluate several backbones, including ModernBERT~\citep{modernbert}, Qwen2.5-1.5B~\citep{qwen2026qwen35}, and Llama-3.2-1B~\citep{llama3}. Across these settings, the learned reward models reach only around 0.3 Spearman correlation with human ratings. This is comparable to the stronger LLM judges in Appendix Table~\ref{tab:llm-judge-personalization}, such as Claude-S4.6 and GPT-5.4, but remains far from a reliable judge of personalization quality.

This difficulty also reflects a broader limitation of evaluating personalization through aggregate judgments. Personalization preference is inherently subjective: even trained human annotators show only moderate agreement with one another. Ideally, personalized response quality should ultimately be judged by the user being personalized for. Our results provide a first step toward showing that current LLM judges and reward models do not fully capture aggregate human preferences, while also suggesting that future work may need to consider user-specific or preference-adaptive reward models rather than a single global judge.

\section{Conclusion}
\label{sec:conclusion}

We study LLM personalization through three stages: attribute extraction, relevance matching, and personalized generation. Across all stages, real human conversations and judgments reveal limitations that synthetic data and LLM-based evaluation can obscure. Models infer noisy attributes, over-select relevant attributes, and often fail to produce responses that humans prefer.

\section{Limitations}

Our study has several limitations. First, we aggregate human annotations into consensus labels or average ratings for simplicity and for compatibility with standard evaluation and training objectives. This allows us to study broad gaps between humans and models, but it also collapses meaningful variation across individual annotators. Personalization is inherently subjective, and different users may reasonably prefer different forms of adaptation, levels of explicitness, or tones. Future work should model such variation more directly, for example through user-specific preference models or evaluation protocols that preserve disagreement rather than treating it only as noise.

Second, our data and annotations are primarily grounded in English-language interactions and likely reflect mostly Western conversational norms. However, personalization is shaped by cultural values, social expectations, and communication styles. What counts as helpful, intrusive, polite, or appropriately personalized may differ substantially across languages and cultural contexts. Extending this framework to multilingual and cross-cultural settings is therefore an important next step.

Third, our study evaluates an initial personalization pipeline based on extracting stable user attributes, selecting relevant attributes, and generating a personalized response. This abstraction is useful for diagnosis, but it does not cover all aspects of real deployed personalization systems. For example, we do not study how memories should be updated over time, how outdated or conflicting attributes should be handled, or how users should control what information is retained and used. These issues are central to building trustworthy long-term personalization systems.

Finally, our training-based interventions are evaluated on the human annotations collected in this study. While they show that human annotations can improve attribute verification and relevance selection, our experiments do not establish that the learned models will generalize to all user populations, domains, or personalization settings. Larger and more diverse human-centered datasets are needed to determine how robust these interventions are beyond our evaluation setting.

\section{Ethical Considerations}

This work uses real user conversations, which raises privacy and consent concerns. Although we rely on an existing public dataset and analyze aggregate model behavior, real conversations may still contain sensitive or context-dependent information. We therefore avoid identifying users and emphasize that personalization systems should not treat inferred attributes as reliable memories without verification and user control.

Our results also show that models can overgeneralize from limited evidence and use personal attributes in ways that feel intrusive or presumptuous. Personalization systems should therefore evaluate not only whether user information is used, but whether it is used accurately, respectfully, and only when it meaningfully improves the response.

Finally, our human annotation tasks require subjective judgments about other users' preferences and may expose annotators to personal content. These judgments are useful but imperfect, and future systems should give users direct control over what is remembered, when it is used, and how it shapes responses.

\section*{Acknowledgments}
We thank all members of the \href{https://languageinteraction.github.io/}{Language Interaction Lab} for their extensive feedback and suggestions on this work. We also thank Shuhaib Mehri for helpful discussions during the early stage of this project. We are also grateful to our Prolific annotators for their careful judgments and contributions to our study. This work used computational resources and services at the \href{https://delta.ncsa.illinois.edu/}{National Center for Supercomputing Applications (NCSA) Delta GPU Cluster} through allocation CIS251116 from the \href{https://access-ci.org/}{Advanced Cyberinfrastructure Coordination Ecosystem: Services \& Support (ACCESS)} program, which is supported by U.S. National Science Foundation grants \#2138259, \#2138286, \#2138307, \#2137603, and \#2138296.

\bibliography{custom}

\clearpage

\appendix

\section{WildChat Preprocessing Details}
\label{app:wildchat}

We build our real-user corpus from WildChat~\citep{zhao2024wildchat}, which contains 1,039,785 conversations. Our preprocessing pipeline applies the following filters.

\paragraph{Language filtering.}
We first retain conversations labeled as English in WildChat's metadata. We then apply an LLM-based classifier (Gemma-4-31B) to verify each conversation is entirely in English, excluding conversations with non-English content such as translation tasks where foreign text appears in user turns (Appendix~\ref{prompt:language-detection}).

\paragraph{User clustering and template removal.}
We group conversations by hashed IP address to form per-user histories. A sliding-window template detector computes 120-character window hashes (stride 60) over normalized user turns. A user is flagged as scripted if any hash window appears in five or more turns with coverage above 40\%, or in eight or more turns absolutely. This removes 5,777 scripted accounts, leaving 92,557 users.

\paragraph{Activity threshold.}
We retain only users with at least 3 source conversations, more than 15 total user messages across all conversations, yielding 16,573 active user histories.

\paragraph{Genuine personal assistant usage.}
We score each user's conversations using an LLM-based classifier that judges how much the user treats the model as a conversational personal assistant rather than as an API endpoint or a jailbreak target (Appendix~\ref{prompt:assistant-usage}). The classifier outputs a score in $[0, 1]$. Users whose conversations score below 0.6 are excluded during attribute extraction.

\section{Attribute Extraction Implementation}
\label{app:attr-extraction}

\paragraph{Model and setup.}
We use Llama-3.3-70B~\citep{llama3} served via vLLM, processing each conversation independently at temperature 0.8 with a maximum of 4,096 output tokens. Conversations exceeding 50,000 characters are split into chunks of up to 50,000 characters; attributes extracted from all chunks are pooled before deduplication. The extraction prompt is in Appendix~\ref{prompt:attr-extraction}.

\paragraph{Post-processing.}
Attributes with confidence below 0.4 are discarded. These confidence values are self-reported by the extraction model and serve only as a coarse first-pass filter. More robust estimates of output reliability such as self-consistency \citep{liu-etal-2026-taming} might further improve this step, which we leave to future work. The remaining attributes are embedded with \texttt{all-MiniLM-L6-v2}~\citep{reimers-gurevych-2019-sentence} and clustered per user using agglomerative clustering (average linkage). We manually inspected four cosine similarity thresholds (0.6, 0.7, 0.8, 0.9) on a held-out sample of users; 0.7 offered the best balance, leaving almost no duplicate attributes while only occasionally merging two genuinely distinct ones. Since missing an attribute is less harmful to downstream tasks than retaining noisy duplicates, this tradeoff was acceptable. Within each cluster, the attribute whose embedding is closest to the confidence-weighted centroid is selected as the representative, and cluster confidence is aggregated using an independence-aware formula that discounts correlated evidence sources.

\section{User Diversity Sampling Details}
\label{app:diversity-sampling}

We select a diverse subset using the following procedure.

\paragraph{Intra-user Diversity.}
For each eligible user, we compute a \emph{generalist score} as the mean pairwise cosine distance among a random subsample of $K = 17$ merged-attribute embeddings (set to the p50 of eligible users' attribute counts). This captures semantic breadth: a user with attributes spanning diverse topics has a high mean pairwise distance, while a specialist user has a low score.

\paragraph{Inter-user farthest-point diversity sampling.}
We divide the eligible pool into five equal-frequency quintile bins on the generalist score. Within each bin, we apply greedy farthest-point sampling on per-user mean attribute embeddings to maximize semantic diversity within each spectrum band. Slots are allocated proportionally across bins, ensuring the sample spans the full specialist-to-generalist spectrum.

\paragraph{Examples.}
Table~\ref{tab:wildchat-user-examples} shows five representative users from the sampled pool, one drawn from each diversity quintile.

\begin{table*}[ht]
\centering
\small
\renewcommand{\arraystretch}{1.4}
\setlength{\tabcolsep}{7pt}
\newcommand{\gsbar}[1]{{\color{gray!30}\rule{2.2cm}{5pt}}\llap{{\color{teal!70!black}\rule{#1\dimexpr2.2cm}{5pt}}}}
\caption{%
  Representative attributes and intra-user diversity scores for five sampled WildChat users,
  spanning the full specialist--generalist spectrum.
  Intra-user Diversity is mean pairwise cosine distance among a user's merged attributes (min-max
  normalised to $[0,1]$; $0=$ specialist, $1=$ generalist).
}
\label{tab:wildchat-user-examples}
\begin{tabularx}{\linewidth}{lc>{\raggedright\arraybackslash}X}
\toprule
\textbf{Profile} & \textbf{Intra-user Diversity} & \textbf{Top-5 Representative Attributes} \\
\midrule
Creative Artist & \gsbar{0.24} \quad 0.24 & Values human creativity and artistic vision over AI-generated art.\ \textbullet\ Interested in omegaverse fanfiction.\ \textbullet\ An artist or illustrator.\ \textbullet\ Skeptical about the benefits of AI in creative fields.\ \textbullet\ Likely a musician or has a strong interest in music. \\
\addlinespace[2pt]
Social Entrepreneur & \gsbar{0.41} \quad 0.41 & Interested in environmental sustainability.\ \textbullet\ Involved with or supports Msitu Africa.\ \textbullet\ Interested in social innovation and its applications.\ \textbullet\ Focused on empowering farmers in Kenya and Africa.\ \textbullet\ May have a professional or entrepreneurial interest in Africa. \\
\addlinespace[2pt]
Software Developer & \gsbar{0.49} \quad 0.49 & A developer familiar with Flutter.\ \textbullet\ Working with XML or docx files.\ \textbullet\ Familiar with Python programming language.\ \textbullet\ A programmer working with XML or HTML documents.\ \textbullet\ Trying to extract text from docx files. \\
\addlinespace[2pt]
Parent / News Consumer & \gsbar{0.55} \quad 0.55 & Interested in Tina Turner's music.\ \textbullet\ Concerned about their child's online safety.\ \textbullet\ Interested in earthquake information, specifically about Turkey.\ \textbullet\ A parent or guardian.\ \textbullet\ May not be aware of recent news or updates about Tina Turner. \\
\addlinespace[2pt]
Generalist & \gsbar{0.68} \quad 0.68 & Values diversity and inclusiveness.\ \textbullet\ Has health considerations, including asthma and possibly psoriasis.\ \textbullet\ Prefers drama movies.\ \textbullet\ A fan of Christopher Nolan's work.\ \textbullet\ Concerned about being targeted for vandalism or theft due to vehicle appearance. \\
\addlinespace[2pt]
\bottomrule
\end{tabularx}
\end{table*}

\section{Uncertain Attribute Categorization}
\label{app:attr-uncertainty}

From our attribute quality annotation (Section~\ref{sec:attributeNoise}), human annotators flagged 1,225 unique attributes as uncertain or rejected (868 are from WildChat, 134 from PersonaLens, 133 from CUPID, 90 from PrefEval). We applied a two-stage GPT-5.4 pipeline to categorize the dominant reason for each flag: Stage 1 infers a free-form reason from the annotation context (Appendix~\ref{prompt:attr-infer}); Stage 2 assigns one of six categories (Appendix~\ref{prompt:attr-classify}). Table~\ref{tab:uncertainty} shows the distribution with representative examples.

\begin{table}[h]
\centering
\small
\setlength{\tabcolsep}{4pt}
\renewcommand{\arraystretch}{1.1}
\caption{Failure modes among 1,225 uncertain or rejected attributes, categorized using GPT-5.4.}
\label{tab:uncertainty}
\resizebox{\columnwidth}{!}{%
\begin{tabular}{lrp{6.2cm}}
\toprule
\textbf{Category} & \textbf{\%} & \textbf{Definition} \\
\midrule
Overgeneralization & 53.9 & Broad or stable user traits are inferred from narrow, one-off, or situational evidence. \\
Missing evidence & 20.3 & The attribute relies on details that are not visible, mismatched, or insufficiently supported by the excerpt. \\
Task-context confusion & 16.1 & The attribute is inferred from the task subject, fictional framing, or assistant-added content rather than the user's own statements. \\
Attribute not standalone & 6.0 & The attribute mainly describes the immediate task or artifact and does not stand alone as a context-independent user trait. \\
Others & 3.7 & -- \\
\bottomrule
\end{tabular}
}
\end{table}

\section{RoBERTa Attribute Verifier: Training Details}
\label{app:roberta-task4}

We fine-tune \texttt{roberta-base}~\citep{liu2020roberta} on the Task~4 attribute quality annotations to classify each extracted attribute as accepted or rejected.

\paragraph{Input format.}
Each example is encoded as \texttt{[CLS] \{conversation snippet\} [SEP] \{attribute\} [SEP]}, truncated to 512 tokens. The snippet is the evidence excerpt shown to human annotators.

\paragraph{Training setup.}
We train for 10 epochs with batch size 16 and learning rate $2 \times 10^{-5}$. We use weighted cross-entropy loss with $F_\beta$ weighting ($\beta = 2.0$) to prioritize recall over precision, since unsupported attributes missed by the verifier are more harmful downstream than false positives. Data is split at the user level to prevent attribute-level contamination across train / validation / test sets.

\section{Relevance Alignment: Training Details}
\label{app:training-task5}

\paragraph{RoBERTa classifier.}
We fine-tune \texttt{roberta-base} on the Task~5 human relevance annotations. Each example is encoded as \texttt{[CLS] \{prompt\} [SEP] \{attribute\} [SEP]}, truncated to 256 tokens. Training uses 10 epochs, batch size 16, and learning rate $2 \times 10^{-5}$, with a prompt-level split: train ($\sim$76\% of prompts), validation (prompts 38--42), and test (prompts 33--37).

\paragraph{GRPO.}
We optimize Qwen3-4B using GRPO via the \texttt{verl} framework~\citep{sheng2024hybridflow} on 4$\times$ A100 80\,GB GPUs. Training runs for 15 epochs with batch size 128, actor learning rate $1 \times 10^{-6}$, and KL loss coefficient 0.001. For each prompt, 5 rollout responses are sampled ($n=5$); the reward function is derived from the human majority-vote relevance label. Maximum prompt length is 1,024 tokens and maximum response length is 4,096 tokens. We optimize the model solely for relevance matching and do not attempt to preserve its performance on other tasks \citep{liu2026mixsd}.

\section{Human Annotation Details}
\label{app:human-annot}

\begin{figure*}[h]
    \centering
    \includegraphics[width=0.85\textwidth]{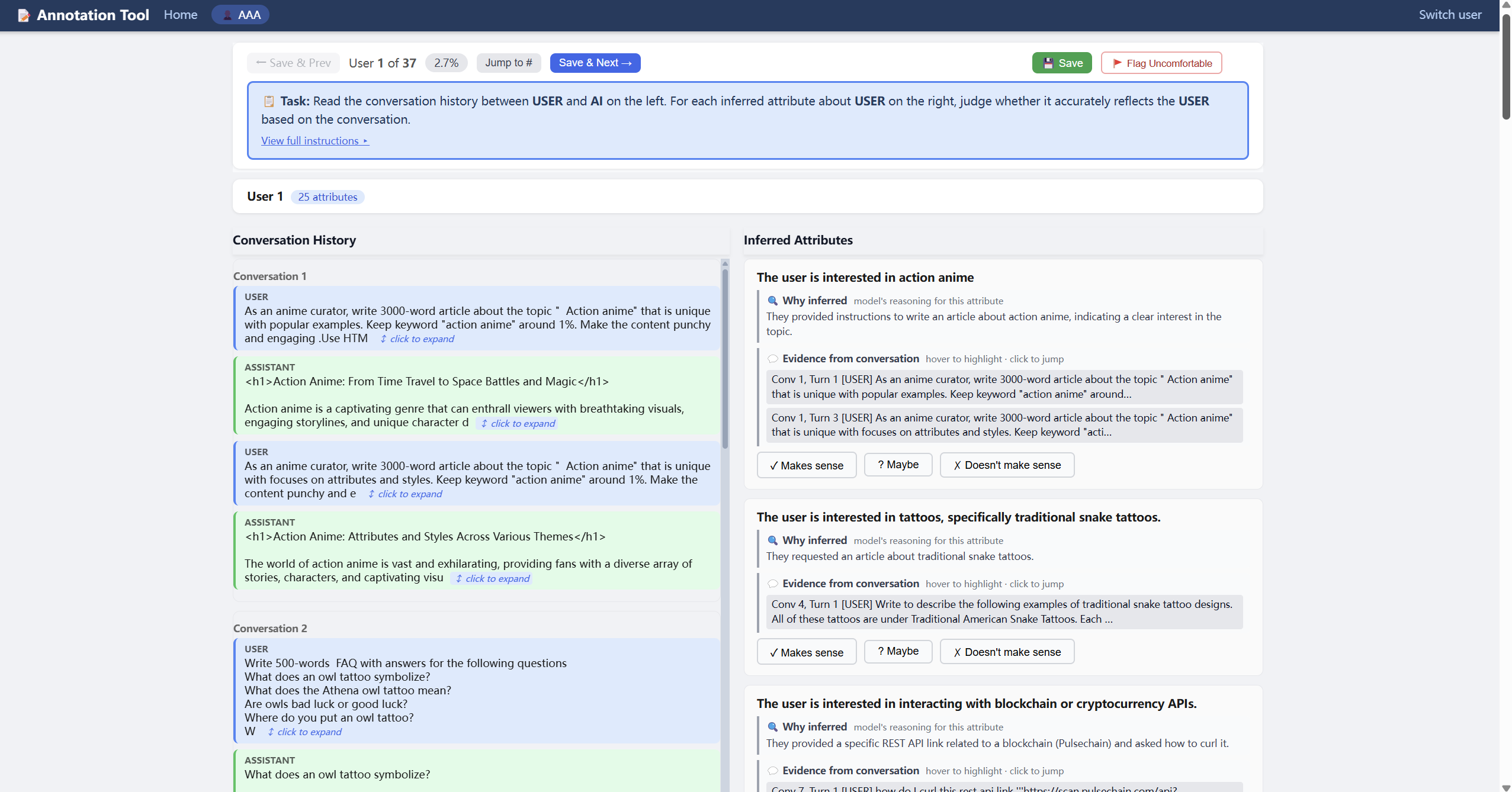}
    \caption{Task 1 annotation interface. Annotators judge each extracted attribute against the user's conversation history using a three-way label.}
    \label{fig:annot-task4}
\end{figure*}

\begin{figure*}[h]
    \centering
    \includegraphics[width=0.85\textwidth]{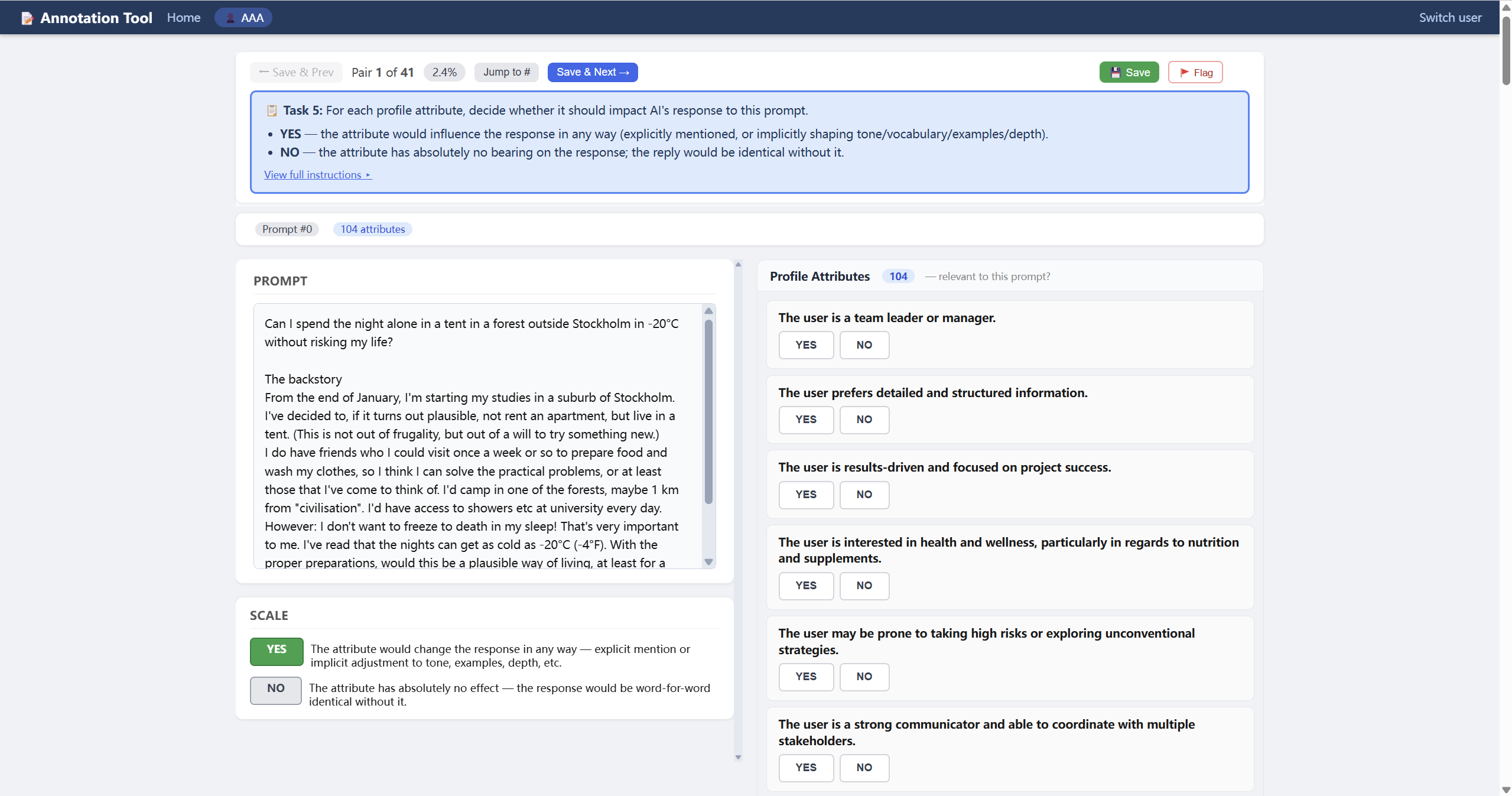}
    \caption{Task 2 annotation interface. Annotators decide YES or NO for each (attribute, prompt) pair, with model votes shown for reference.}
    \label{fig:annot-task5}
\end{figure*}

\begin{figure*}[h]
    \centering
    \includegraphics[width=0.85\textwidth]{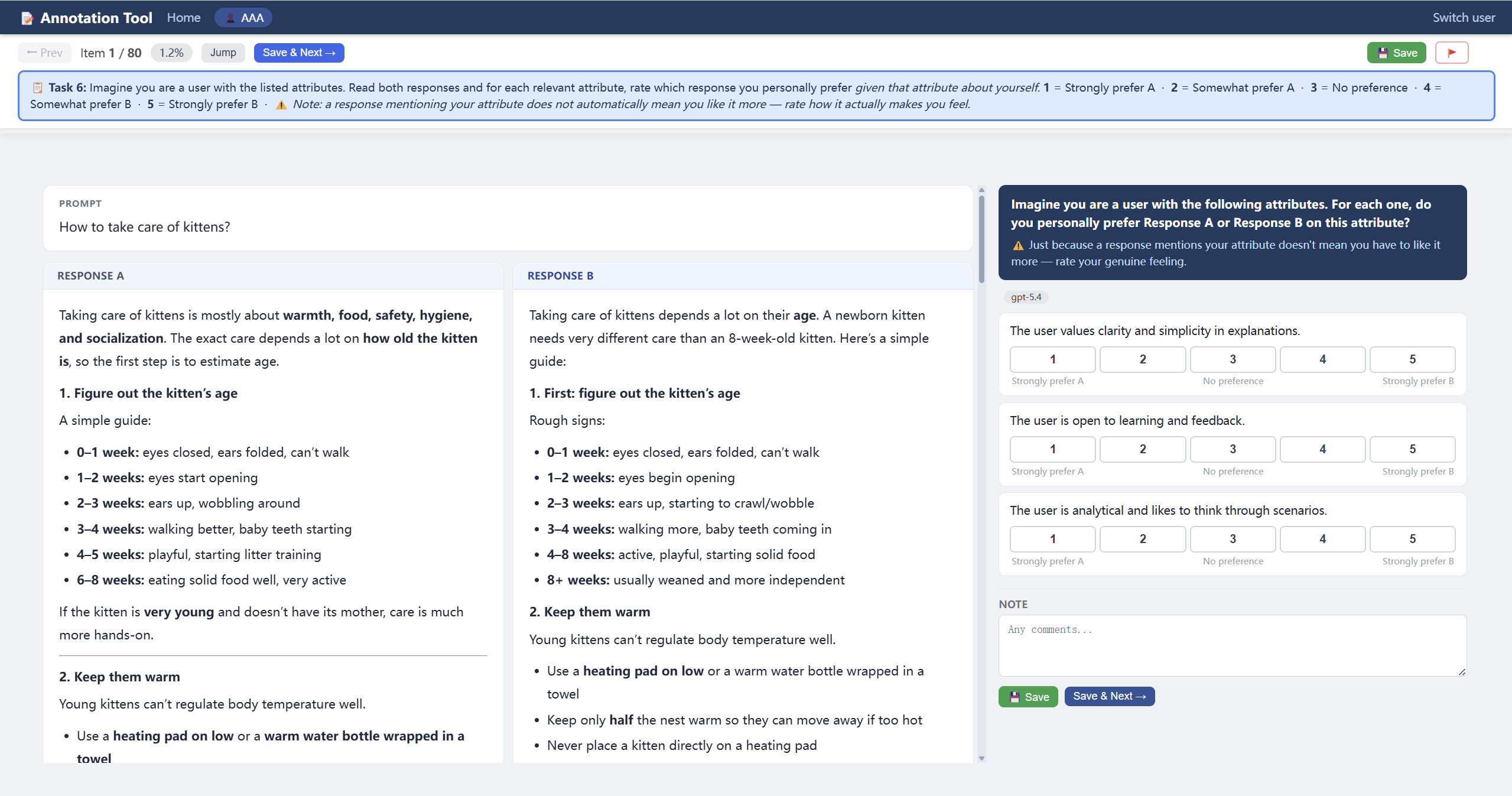}
    \caption{Task 3 annotation interface. Annotators rate attribute-level preference between two anonymized responses on a 1--5 scale.}
    \label{fig:annot-task6}
\end{figure*}

Three human annotators participated in all annotation tasks, selected through a pilot study based on annotation quality and self-consistency. All tasks were served through a custom web-based annotation interface; Screenshots of the interfaces for Tasks 4, 5, and 6 are shown in Figures~\ref{fig:annot-task4}, \ref{fig:annot-task5}, and \ref{fig:annot-task6}, respectively. We recruited annotators through Prolific and paid them at an estimated rate of \$18 per hour, which is above common minimum-wage standards and intended to provide fair compensation for English-fluent crowdworkers. Participants were shown study instructions explaining how their judgments would be used, and provided consent before completing the task.

\paragraph{Stage 1: Attribute Quality.}
Annotators viewed each user's conversation history alongside extracted attributes and supporting evidence snippets. For each attribute, they selected one of three labels: \textit{Makes sense} (plausible given the conversation), \textit{Maybe} (uncertain), or \textit{Doesn't make sense} (unsupported or incorrect), and could flag uncomfortable instances. A total of 77 users were annotated, yielding 1,983 (item, attribute) judgments across three annotators.

\paragraph{Stage 2: Relevance Selection.}
Annotators viewed each (attribute, prompt) pair and selected YES or NO following the criteria in Appendix~\ref{prompt:relevance}. Three annotators each judged all 3,973 (attribute, prompt) pairs. Human annotators mark approximately 20\% of attributes as relevant on average, substantially below the 40--60\% yes-rate observed from LLM judges.

\paragraph{Stage 3: Personalization Quality.}
Annotators were shown a user's attribute profile, a prompt, and two responses (A and B, randomly assigned to no-context and personalized conditions) rendered side by side. For each relevant attribute, annotators rated on a 1--5 scale which response they preferred given that attribute (1 = strongly prefer A; 5 = strongly prefer B). A total of 80 items were annotated, producing 367 (item, attribute) ratings across three annotators.

\begin{table}[t]
\centering
\small
\setlength{\tabcolsep}{4pt}
\renewcommand{\arraystretch}{1.05}
\resizebox{\linewidth}{!}{%
\begin{tabular}{lcc@{\hspace{14pt}}|lcc}
\toprule
\textbf{Judge} & \textbf{Avg. Rating} & \textbf{Spearman $\rho$} &
\textbf{Judge} & \textbf{Avg. Rating} & \textbf{Spearman $\rho$} \\
\midrule
Human & 3.176 & -- &
Llama-3.3-70B & 4.019 & \textbf{0.376} \\
Claude-S4.6 & 3.428 & 0.362 &
Qwen3.5-27B & 3.487 & 0.182 \\
GPT-5.4 & 3.523 & 0.312 &
Gemma-4-31B & 3.409 & 0.111 \\
\bottomrule
\end{tabular}%
}
\caption{Agreement between LLM judges and human average ratings for personalized response quality. Avg. Rating denotes the average preference score, and $\rho$ denotes Spearman correlation with human ratings.}
\label{tab:llm-judge-personalization}
\end{table}

\section{GRPO Relevance Reasoning Example}
\label{app:relevance-example}

Table~\ref{tab:grpo-example} shows a side-by-side comparison of the base Qwen3-4B model and the GRPO-optimized model on the same (attribute, prompt) pair. The base model treats any possible connection between the attribute and the response as sufficient grounds for YES. The GRPO model catches itself mid-reasoning and applies a more calibrated standard.

\begin{table*}[h]
\centering
\small
\setlength{\tabcolsep}{6pt}
\renewcommand{\arraystretch}{1.15}
\begin{tabular}{p{2.6cm} p{5.9cm} p{5.9cm}}
\toprule
 & \textbf{Base Qwen3-4B} & \textbf{GRPO-optimized} \\
 \midrule
\textbf{Prompt} & \multicolumn{2}{p{12.1cm}}{\textit{What is the difference between a mode and a scale in music theory?}} \\
\midrule
\textbf{Attribute} & \multicolumn{2}{p{12.1cm}}{\textit{The user is a creative individual.}} \\[3pt]

\midrule
\textbf{Reasoning} &
Well, a creative person might appreciate more examples or deeper explanations. If the AI is personalized, it might adjust the response to be more detailed, use more examples, or explain in a way that connects to creative processes. The AI could mention how different modes are used in composing or how scales evoke certain moods. Therefore, the attribute would influence the response. $\to$ \textbf{[YES]}
&
Well, if the user is creative, maybe the AI should use more examples or analogies that relate to creativity. \textbf{\textit{But wait, the question is about a basic definition. The answer is factual, so this attribute should not change the factual answer.}} The response would remain the same in terms of content, structure, and accuracy regardless of the user's personality traits. $\to$ \textbf{[NO]} \\

\bottomrule
\end{tabular}
\caption{Base vs.\ GRPO-optimized Qwen3-4B reasoning on the same (attribute, prompt) pair. The base model conflates possible influence with meaningful influence. The GRPO model self-corrects (highlighted) and applies a more conservative, human-aligned standard.}
\label{tab:grpo-example}
\end{table*}

\section{Prompts}
\label{app:prompts}

We are aware of the extensive existing work on prompt optimization and acknowledge that prompt quality can affect model performance~\citep{pryzant-etal-2023-automatic, zhang2026sprig, khattab2024dspy, zhang2025crosslingualpromptsteerabilityaccurate}. However, it is not feasible to test all possible prompt settings, nor is this the main focus of our paper. Instead, we make every effort to ensure that our comparisons are as fair as possible and that the prompts we evaluate have been iterated sufficiently, such that further simple improvements are unlikely to yield substantially higher scores. This helps ensure that our conclusions are generalizable and convincing.

\subsection{Language Detection}
\label{prompt:language-detection}

Used to verify each conversation is entirely in English during WildChat preprocessing (Appendix~\ref{app:wildchat}).

{\tiny
\begin{lstlisting}
You are a language detection classifier. Your task
is to determine whether a conversation contains
meaningful non-English language usage.

Definition:
Mark is_english as false ONLY if there is clear,
natural-language usage of a non-English language
(e.g., full words, phrases, or sentences in another
language).

Do NOT count the following as non-English:
- Programming code, syntax, or identifiers
  (e.g., SQL, Python, function names)
- Common loanwords used in English
  (e.g., resume, cafe, naive)
- Proper nouns or names (e.g., Beyonce, Pokemon)
- Acronyms or abbreviations (e.g., LLM, GPU, API)
- Mathematical notation, LaTeX, or symbols
- Isolated foreign words widely adopted in English
  that do not form a sentence

Mark is_english as false ONLY if:
- A full or partial sentence is written in another
  language
- Multiple words form a coherent non-English phrase
- The user or assistant is clearly communicating in
  another language
- A translation request includes actual non-English
  text

Otherwise, mark is_english as true.

Respond with a JSON object in this exact format:
{"is_english": true} or {"is_english": false}

Conversation excerpt:
{transcript}

JSON response:
\end{lstlisting}
}

\subsection{Assistant-Like Usage Scoring}
\label{prompt:assistant-usage}

Used to score how much a user treats the model as a personal assistant during WildChat preprocessing (Appendix~\ref{app:wildchat}).

{\tiny
\begin{lstlisting}
You are given several conversations between a USER
and an AI model.

Your task is to judge how much the USER treats the
AI as a conversational assistant, as opposed to:
- using it purely as an API endpoint for repeated,
  templated, similar calls,
- using it mainly for explicit/toxic content without
  seeking real help,
- or trying to attack, exploit, or jailbreak the
  model.

The following behavior of USER should STRONGLY
DECREASE the score and usually indicates
NON-assistant usage (score typically below 0.4):
- Repeatedly sends highly similar templated prompts
  across many conversations.
- Focuses on explicit or toxic content with no
  genuine information-seeking intent.
- Only attempts to bypass safety or exploit the
  model.

IMPORTANT: You must base your judgment ONLY on the
USER messages, and MUST NOT use the ASSISTANT
replies as evidence for scoring.

Output format (STRICT):
You MUST output ONLY a valid JSON object:
{
  "reasoning": "your analysis",
  "score": 0.0
}

Rules:
- "score" must be a number between 0 and 1.
- Do NOT output any text outside the JSON.
- The entire output must be parseable by json.loads.

Here are the conversations:
{transcript}
\end{lstlisting}
}

\subsection{Attribute Extraction}
\label{prompt:attr-extraction}

Used to extract personalizable user attributes from conversation transcripts (Appendix~\ref{app:attr-extraction}).

{\tiny
\begin{lstlisting}
You are analyzing a multi-turn conversation to infer
personalizable attributes about the USER.

Your output must be a single JSON object:
{"attributes": [{"attribute": "<attribute>",
  "reason": "<reason>", "confidence": <confidence>}]}

Where:
- <attribute>: a concise statement describing
  something personal about the USER (preference,
  habit, background, goal, or communication style).
- <reason>: a short explanation of why you inferred
  this attribute from the USER's words or behavior.
- <confidence>: a float between 0.0 and 1.0.

Guidelines:
1. Use only USER messages for reasoning. Ignore
   ASSISTANT messages as evidence.
2. Inference is encouraged even for implied
   attributes, but use lower confidence.
3. Each attribute should describe a stable, general
   aspect of the USER, not a one-time statement.
4. Confidence scale:
   0.9-1.0: Explicitly stated or very clear.
   0.6-0.89: Clearly implied or repeated hints.
   0.3-0.59: Weakly implied or uncertain.
   0.1-0.29: Very speculative but plausible.
5. Be exhaustive. Include as many distinct
   personalizable attributes as possible.

Return only the JSON object. Do not include any
explanation or comments.

Conversation:
{transcript}
\end{lstlisting}
}

\subsection{Uncertain Attribute Inference}
\label{prompt:attr-infer}

Stage 1 of uncertain attribute categorization: GPT-5.4 infers a free-form reason for each non-OK annotator judgment (Appendix~\ref{app:attr-uncertainty}).

{\tiny
\begin{lstlisting}
You are analyzing annotation data from a user
attribute quality study.

Annotators read conversations between a USER and an
AI assistant, then judged each extracted attribute:
- "Makes sense" -- accurately reflects the user
- "Maybe" -- uncertain based on the conversation
- "Doesn't make sense" -- incorrect or unsupported

The attribute below was marked "{dominant_judgment}"
by annotators. Infer the most likely reason a
careful annotator would give this rating.

## Attribute
{attribute}

## Extraction rationale
{reason}

## Conversation excerpt shown to annotators
(Evidence turns highlighted; neighboring turns shown
for context.)
{conv_excerpt}

## Instructions
Give a free-form analysis of why this attribute may
have received a non-OK judgment.
Do NOT force the reason into a predefined category.

Possible issues you may consider include, but are
not limited to:
- The attribute may overgeneralize beyond what the
  conversation supports.
- The attribute may depend on a temporary task
  context rather than a stable user trait.
- The attribute may combine multiple weak clues into
  a stronger claim than warranted.
- The attribute may not be standalone.
- The extraction rationale may not actually support
  the attribute.
- The evidence may be too thin, ambiguous, or
  contradicted by nearby turns.

Respond with ONLY a JSON object:
{
  "inferred_reason": "<2-4 sentences>",
  "evidence_assessment": "<brief assessment>"
}
\end{lstlisting}
}

\subsection{Uncertain Attribute Classification}
\label{prompt:attr-classify}

Stage 2 of uncertain attribute categorization: GPT-5.4 assigns each inferred reason to one of six canonical categories (Appendix~\ref{app:attr-uncertainty}).

{\tiny
\begin{lstlisting}
You are building an error taxonomy for LLM-extracted
user attributes.

Below is a candidate category list. These categories
are provisional. Use one if it fits well, but you
may propose a new category when none of the
candidates captures the reason clearly.

## Candidate categories
{candidate_categories}

## Cases to classify
{entries}

## Instructions
For each case:
1. Assign exactly one category.
2. Prefer an existing candidate category when it
   clearly fits.
3. Create a new category only when the existing
   candidates would blur an important distinction.
4. Keep new category names short, general, reusable.
5. Do not create near-duplicates of existing
   categories.

The two seed categories are especially important:
- overgeneralization: the attribute makes a broader,
  stronger, or more stable claim than the evidence
  supports.
- attribute-not-standalone: the attribute cannot be
  judged as a self-contained user attribute because
  it depends on missing context, a specific quoted
  object, another attribute, or a narrow task.

Respond with ONLY a JSON object:
{
  "classifications": [
    {
      "case_id": <int>,
      "category": "<category-name>",
      "is_new_category": <true/false>,
      "category_description": "<one sentence>",
      "classification_reason": "<one sentence>"
    }
  ],
  "new_categories": [
    {
      "name": "<new-category-name>",
      "description": "<one sentence>",
      "merge_with_existing": "<or empty string>"
    }
  ]
}
\end{lstlisting}
}

\subsection{Attribute Verifier: Zero-Shot}
\label{prompt:attr-verifier-basic}

The baseline verifier prompt (Table~\ref{tab:attribute-verifier}, ``zero-shot'' rows) mirrors the instruction given to human annotators in Task 4.

{\tiny
\begin{lstlisting}
Task: Read the conversation history between USER
and AI. For each inferred attribute about USER,
judge whether it accurately reflects the USER based
on the conversation.

## Conversation History
{snippet}

## Inferred Attribute
{attribute}

## Explanation Given for the Attribute
{reason}

Give the final answer as \boxed{accepted} or
\boxed{rejected}.
\end{lstlisting}
}

\subsection{Attribute Verifier: Optimized}
\label{prompt:attr-verifier-optimized}

Used as the optimized LLM verification prompt for detecting unsupported extracted attributes (Table~\ref{tab:attribute-verifier}). Judgment criteria are derived from the failure modes identified in Section~\ref{sec:attributeNoise}.

{\tiny
\begin{lstlisting}
## Task

Read the conversation history between USER and AI.
Then judge whether the inferred attribute about USER
accurately reflects the USER based on the
conversation.

## Conversation History

{snippet}

## Inferred Attribute

{attribute}

## Explanation Given for the Attribute

{reason}

## Judgment Criteria

Mark the attribute as ACCEPTED if:
- The conversation provides clear evidence for this
  attribute.
- The attribute accurately describes the USER, not
  just the AI or the current task.
- The attribute is specific enough and does not
  exaggerate what the conversation shows.

Mark the attribute as REJECTED if:
- The conversation does not provide enough evidence.
- The attribute is only a guess or a possible
  interpretation.
- The attribute over-generalizes from one isolated
  request.
- The attribute describes only the current task, not
  a reusable fact about the USER.
- The explanation adds assumptions that are not shown
  in the conversation.

## Instructions

Briefly explain your reasoning, then give the final
answer. Use exactly one of the following labels:

\boxed{accepted}
\boxed{rejected}
\end{lstlisting}
}

\subsection{Attribute Relevance Annotation}
\label{prompt:relevance}

Used for both LLM judges and human annotators in the relevance selection task (Appendix~\ref{app:human-annot}).

{\tiny
\begin{lstlisting}
You are helping evaluate a personalized AI assistant.

Task: Given a prompt sent by a user and an attribute
from the user's profile, decide whether the attribute
should be taken into consideration when personalizing
the AI's response.

Criteria:
- Answer YES if the attribute would change the
  response in ANY way compared to a generic reply,
  whether explicitly (directly mentioning the
  attribute) or implicitly (adjusting tone,
  vocabulary, examples, level of detail, or framing).
- Answer NO if the attribute has absolutely no effect
  on how a thoughtful AI would respond.

You MUST end your response with a line containing
only:
Answer: YES
or
Answer: NO

User Attribute: {attribute}
Prompt: {prompt}
\end{lstlisting}
}

\section{Use of Large Language Models}
We acknowledge that we only used LLMs to check grammatical errors in the paper and to improve the clarity of expression.

\section{Licenses}
All data and code will be publicly released under the CC BY-SA 4.0 license. We manually inspected the data for personally identifying or offensive content, avoided reporting any user-identifying examples, and analyzed results only in aggregate to reduce privacy and exposure risks.

\end{document}